\pdfoutput=1

\documentclass[11pt]{article}

\usepackage{ACL2023}

\usepackage{times}
\usepackage{latexsym}
\usepackage{amsmath}
\usepackage{amsfonts}
\usepackage{bm}
\usepackage{amssymb}
\usepackage{graphicx}
\usepackage{caption}
\usepackage{booktabs}
\usepackage{multicol}
\usepackage{multirow}
\usepackage{subcaption}
\usepackage[ruled,linesnumbered]{algorithm2e}
\usepackage{xcolor}
\usepackage{colortbl}
\usepackage[T1]{fontenc}

\usepackage[utf8]{inputenc}

\usepackage{microtype}

\usepackage{inconsolata}
\newcommand{\bA}{\mathbf{A}}

\newcommand{\bS}{\mathbf{S}}

\newcommand{\bD}{\mathbf{D}}
\newcommand{\bW}{\mathbf{W}}

\SetCommentSty{mycommfont}

\title{An AMR-based Link Prediction Approach for Document-level Event Argument Extraction}

\author{
Yuqing Yang\textsuperscript{1}\thanks{\ \ Work done during internship at Amazon Shanghai AI Lab.} \thanks{\ \ Equal contribution.} , 
Qipeng Guo\textsuperscript{2}\footnotemark[2] , 
Xiangkun Hu\textsuperscript{2}, 
Yue Zhang\textsuperscript{3}, 
Xipeng Qiu\textsuperscript{1}\thanks{\ \ Corresponding author.} , 
Zheng Zhang\textsuperscript{2}  \\
\textsuperscript{1}School of Computer Science, Fudan University \\
\textsuperscript{2}Amazon AWS AI, \textsuperscript{3}School of Engineering, Westlake University \\
\texttt{yuqingyang21@m.fudan.edu.cn}, \texttt{\{gqipeng, xiangkhu, zhaz\}@amazon.com} \\
\texttt{xpqiu@fudan.edu.cn}, \texttt{zhangyue@westlake.edu.cn} \\}

\begin{document}
\maketitle
\begin{abstract}
Recent works have introduced Abstract Meaning Representation (AMR) for Document-level Event Argument Extraction (Doc-level EAE), since AMR provides a useful interpretation of complex semantic structures and helps to capture long-distance dependency. However, in these works AMR is used only implicitly, for instance, as additional features or training signals. Motivated by the fact that all event structures can be inferred from AMR, this work reformulates EAE as a link prediction problem on AMR graphs. 

Since AMR is a generic structure and does not perfectly suit EAE, we propose a novel graph structure, Tailored AMR Graph (TAG), which compresses less informative subgraphs and edge types, integrates span information, and highlights surrounding events in the same document. With TAG, we further propose a novel method using graph neural networks as a link prediction model to find event arguments. 

Our extensive experiments on WikiEvents and RAMS show that this simpler approach outperforms the state-of-the-art models by 3.63pt and 2.33pt F1, respectively, and do so with reduced 56\% inference time. The code is availabel at \url{https://github.com/ayyyq/TARA}.
\end{abstract}

\section{Introduction}
Event Argument Extraction (EAE) is a long-standing information extraction task to extract event structures composed of arguments from unstructured text \citep{DBLP:journals/access/XiangW19}. Event structures can serve as an intermediate semantic representation and be further used for improving downstream tasks, including machine reading comprehension \citep{DBLP:conf/emnlp/HanHSBNRP21}, question answering \citep{DBLP:conf/cikm/CostaGD20}, dialog system \citep{DBLP:conf/aime/ZhangCB20}, and recommendation system \citep{DBLP:conf/acl/LiZLPWCWJCVNF20}. Despite the large performance boost by Pre-trained Language Models (PLMs), extracting complex event structures across sentences is still challenging \citep{DBLP:conf/acl/EbnerXCRD20}. 

\begin{figure}[t]
    \centering
    \includegraphics[width=0.9\linewidth]{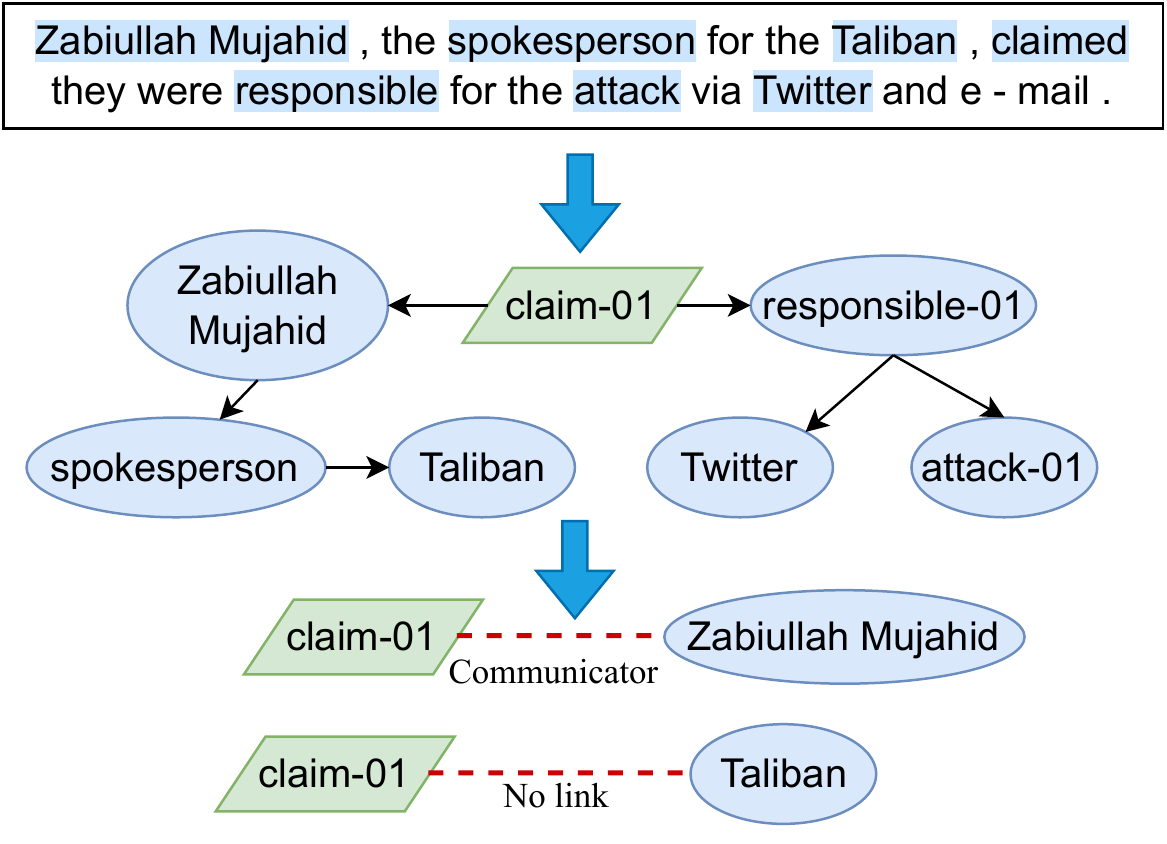}
    \caption{Treating EAE as a link prediction problem, where the green box means the event trigger and blue circles are non-trigger nodes. The highlighted text means they are captured by the graph. We parse the document to a tailored AMR graph and apply a GNN model to predict edges between the trigger and other nodes. In this example, the model predicts one argument, ``Zabiullah Mujahid'', with the role of ``Communicator''. }
    \label{fig:intro-link}
\end{figure}


In real-world text, event structures are usually distributed in multiple sentences \citep{DBLP:conf/naacl/LiJH21}. To capture cross-sentence and multi-hop structures, \citet{DBLP:conf/naacl/XuWLZCS22} introduces Abstract Meaning Representation (AMR) graphs to assist the model in understanding the document. Their main idea is to take AMR as additional features to enrich span representations. \citet{DBLP:journals/corr/abs-2205-12490} and \citet{DBLP:conf/acl/WangW0L000L020} utilize AMR graphs to provide training signals via self-training and contrastive learning, respectively. These methods exemplify that introducing AMR information facilitates the model's understanding of complex event structures. However, previous works implicitly use AMR information by enriching neural sequential models rather than making explicit use of discrete structures. Intuitively, discrete AMR structures can force the model to better focus on predicate-argument structures and the content most related to EAE, therefore having stronger effect than implicit AMR.

We aim to exploit the potentials of explicit AMR for improving EAE by formulating EAE as a link prediction task, and Figure~\ref{fig:intro-link} illustrates the framework. We parse the input document to a graph structure and adopt a link prediction model to find event arguments. We determine if a node is an argument by whether it is connected to the trigger node or not. 
The advantages of formulating EAE as a link prediction problem are three-fold: 1) 
AMR graph is typically more compact than raw text (see Sec-\ref{sec:tAMR}), so processing AMR to find arguments would be simple and efficient. 2) Dependencies among multiple arguments and events are explicitly captured, while previous works \citep{DBLP:conf/acl/LiaoG10, DBLP:conf/acl/DuLJ22} have pointed out the importance of these dependencies which are only implicitly considered in the feature space. 3) The simpler model architecture and sparse graphs can lead to improvement over efficiency, as our experiments show (up to 56\% inference time saving).


The proposed method assumes that AMR graphs contain all necessary information for EAE. However, the original AMR graphs generated by off-the-shelf AMR parsers do not meet this assumption. First, they cover only 72.2\% event arguments in WikiEvents, 
impeding the performance of EAE models directly on the parsed AMR graphs. The primary problem is that AMR graphs are defined at word-level, but an event argument could be a text span.
Second, the Smatch score of SOTA AMR parsers is around 85 \citep{DBLP:conf/acl/00010022}, which causes information loss as well.
To address the above issue, we propose a novel Tailored AMR Graph (TAG), which compresses information irrelevant to EAE, merges words into text spans via a span proposal module, and highlights the surrounding events in the same document to encourage their communication. Particularly, the number of nodes in TAG equals around 47\% of words in WikiEvents, which is a significant reduction. Since too much distracting information is a major challenge of document-level tasks, we also expect performance gains from focusing on TAG, which is evidenced by our experiment results. TAG can cover all EAE samples if the span proposal module adds enough text spans, and we will discuss the trade-off between the recall of spans and model efficiency in Appendix-\ref{sec:choice-of-k}. 




Although there is a large design space for the link prediction model, we choose a simple architecture that stacks GNN layers on top of pre-trained text encoders. The whole model is called TARA for \textbf{T}ailored \textbf{A}M\textbf{R}-based \textbf{A}rgument Extraction.
We conduct extensive experiments on latest document-level EAE datasets, WikiEvents \citep{DBLP:conf/naacl/LiJH21} and RAMS \citep{DBLP:conf/acl/EbnerXCRD20}. TARA achieves 3.63pt and 2.33pt improvements of F1 against the SOTA, respectively. 
Since interactions in GNN are sparse, the computation cost of our model is also lower, saving up to 56\% inference time.

To our knowledge, we are the first to formulate EAE as a link prediction problem on AMR graphs. 





\section{Methodology}
In this section, we first explain how to formulate EAE as a link prediction problem and discuss the benefits of doing so (Sec-\ref{sec:link-pred}). To make AMR graphs better suit the EAE task and ensure the reformulation is lossless, we provide a series of modifications for AMR graphs, resulting in a compact and informative graph, named Tailored AMR Graph (TAG) (Sec-\ref{sec:tAMR}).



\subsection{EAE as Link Prediction}
\label{sec:link-pred}

Formally, given a document $\bD$ and an event trigger $\tau$ with its event type $e$, the goal of Doc-level EAE is to extract a set of event arguments $\bA$ related to $\tau$. We formulate EAE as a link prediction problem, which is defined on TAG. Suppose all nodes in TAG are aligned with text spans in the input sequence, triggers and arguments are captured in the graph, and the node corresponding to the event trigger is marked (we will discuss how to satisfy these in Sec-\ref{sec:tAMR}).


Thus, we apply a link prediction model to the tailored AMR graph $\mathcal{G}_t$ of the document $\bD$. If the model predicts there is an edge connecting a node $u$ and the event trigger $\tau$ with the type $r$, we say the corresponding text span of $u$ is an argument, and it plays the role $r$ in the event with trigger $\tau$. We illustrate this procedure in Figure~\ref{fig:intro-link}, and it also shows the tailored AMR graph removes a large amount of distracting information in the input text. Note that the removed text participates in constructing initial node representations, so the model can still access their information as context. Detailed implementation is shown in Sec-\ref{sec:implementation}. 




\subsection{Tailored AMR Graph for EAE}  
\label{sec:tAMR}

TAG can be built on vanilla AMR graphs generated by an off-the-shelf AMR parser \citep{DBLP:conf/acl/00010022,DBLP:conf/emnlp/AstudilloBNBF20}, which also provides the alignment information between nodes and words. As mentioned above, vanilla AMR graphs are insufficient to solve EAE, so we clean the graph by compressing bloated subgraphs, enrich the graph with span boundary information derived by a span proposal module, and highlight the surrounding events to encourage interactions among multiple events. 

\paragraph{Coalescing edges}
We follow previous works \citep{DBLP:conf/naacl/ZhangJ21,DBLP:conf/naacl/XuWLZCS22} and cluster the fine-grained AMR edge types into main categories as shown in Table~\ref{tab:amr-edge-type} and parse the document sentence by sentence before fully connecting the root nodes of all the sentences.


\begin{figure}[t]
    \centering
    \includegraphics[width=\linewidth]{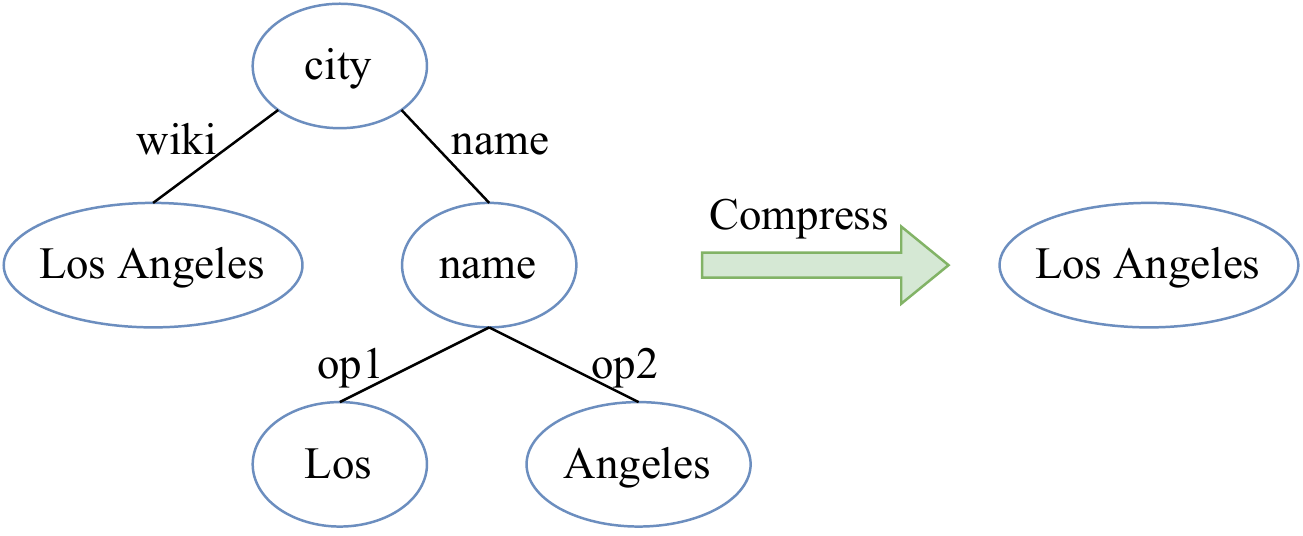}
    \caption{An example of the compression process in Tailored AMR Graph. A subgraph will be replaced by a node with the merged content.}
    \label{fig:method_compress}
\end{figure}

\begin{table}[!tbp]
\centering
\small
\caption{Coalescing edge types of TAG, where each \textit{ARGx} is treated as an individual cluster.}
\begin{tabular}{cc}
    \toprule
    \bf Categories & \bf AMR edge types \\
    \midrule
    Spatial & location, destination, path \\
    \hline
    Temporal & year, time, duration, decade, weekday \\
    \hline
    Means & instrument, manner, topic, medium \\
    \hline
    Modifiers & mod, poss \\
    \hline
    Operators & op-X \\
    \hline
    Prepositions & prep-X \\
    \hline
    Core Roles & ARG0, ARG1, ARG2, ARG3, ARG4 \\
    \hline
    Others & \textit{Other AMR edge types} \\
    \bottomrule
\end{tabular}

\label{tab:amr-edge-type}
\end{table}

\paragraph{Compressing Subgraphs}

AMR is rigorous and tries to reflect all details as much as possible. For example, Figure~\ref{fig:method_compress} shows that a vanilla AMR graph uses five nodes to represent an entity ``\textit{Los Angeles}''. Since EAE does not require such detailed information, we can compress the subgraph to a single node. We find that about 36\% of nodes and 37\% of edges can be removed by compression. Note that all incoming and outgoing edges of the subgraph to be compressed will be inherited, so that the compression does not affect the rest of the graph. A streamlined graph not only improves efficiency and saves memory but also promotes the training of GNN since a larger graph often requires a deeper GNN. The compression procedure only relies on the vanilla AMR graph, so it is a one-time overhead for each sample. The detailed compression rules are described in Appendix-\ref{sec:compress}.  


\begin{figure*}[ht]
    \centering
    \includegraphics[width=0.95\linewidth]{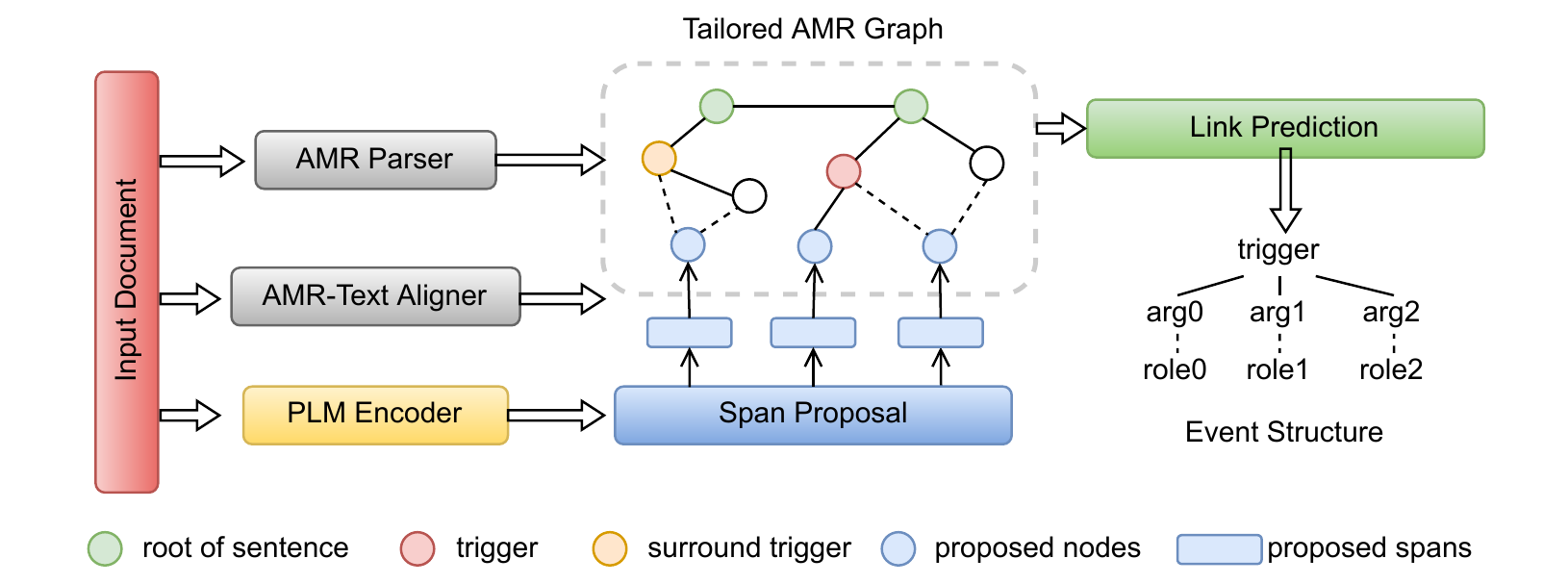}
    \caption{Overview of our method TARA. We adopt an AMR paser and aligner to parse the input document for AMR information, and it is combined with spans generated by a span proposal module to form the Tailored AMR Graph. This graph also contains task-relevant information, which is detailed in Sec-\ref{sec:tAMR}. We further apply a link prediction model on top of the graph to find event arguments. }
    \label{fig:method-main}
\end{figure*}

\paragraph{Missing Spans}
The vanilla AMR graph fails to cover span-form arguments since it is defined at the word level, harming the performance on more than 20\% of EAE samples. 
To overcome this issue, we add the span information $\bS$, which is generated by a span proposal module, 
to $\mathcal{G}_t$ as shown in Figure~\ref{fig:method-main}.
We follow the idea introduced in \citet{DBLP:conf/naacl/ZhangJ21} to merge the generated spans with existing AMR nodes. If a generated span perfectly matches a node's position in the text sequence according to the alignment information, we add a special node-type embedding to the node's initial representation so that the model can know the span proposal module announces this node. If a generated span partially matches a node, we add a new node to represent this span and inherit connectives from the partially matched node. We also add a special edge between this node and the new node to indicate their overlap. If a generated span fails to match any existing nodes, we add a new node and connect it to the nearest nodes to its left and right with a special edge.

\paragraph{Surrounding Events}
Events in a document are not isolated. A recent work \citep{DBLP:conf/acl/DuLJ22} augments the input with the text that contains other events, but the utilization of AMR graphs offers a simpler solution. We add node-type embeddings to indicate that a node is the current trigger or surrounding event triggers in the same document. This modification encourages communication between multiple event structures, and the consistency between event structures can help to extract as many correct arguments as possible. For example, the \textit{Victim} of an \textit{Attack} event is likely to be the \textit{Victim} of a \textit{Die} event, while less likely to be the \textit{Defendant} of an \textit{ChargeIndict} event in the same document.


\subsection{Implementation}
\label{sec:implementation}

We propose a novel model to find event arguments based on TAG, and Figure~\ref{fig:method-main} gives an overview of our method. We first parse the input document with an AMR parser and aligner to obtain the vanilla AMR graph, and coalesce edges and compress subgraphs to preprocess it as described in Sec-\ref{sec:tAMR}. We then enrich the graph with spans generated by a span proposal module. Next, we use token-level features output by a pre-trained text encoder to initialize node representation according to the alignment information. Finally, a GNN-based link prediction model is applied to predict event arguments.
 
\paragraph{Encoder Module}
Given an input document $\bD = \{w_1, w_2, \dots, w_n\}$, we first obtain the contextual representation $\mathbf{h}_i$ for each word $w_i$ using a pre-trained language model such as BERT or RoBERTa:
\begin{displaymath}
\mathbf{H} = [\mathbf{h}_1, \mathbf{h}_2, \dots, \mathbf{h}_n] = \mathrm{PLM}([w_1, w_2 \dots, w_n]).
\end{displaymath}

For a text span $s_{ij}$ ranging from $w_i$ to $w_j$, 
we follow \citet{DBLP:conf/naacl/XuWLZCS22} to calculate its contextual representation $\mathbf{x}_{s_{ij}}$ by concatenating the start representation $\mathbf{h}_i$, the end representation $\mathbf{h}_j$, and the average pooling of hidden states of the span, which would inject span boundary information. Formally,
\begin{displaymath}
\mathbf{x}_{s_{ij}} = \bW_{0}\left[\bW_1 \mathbf{h}_i; \bW_2 \mathbf{h}_j; \frac{1}{j-i+1}\sum_{t=i}^j\mathbf{h}_t\right],
\end{displaymath}
where $\bW_{0}, \bW_{1}, \bW_{2}$ are trainable parameters.

\paragraph{Span Proposal Module}
To find arguments as many as possible, we enumerate all spans up to length $m$. Following \citet{DBLP:conf/acl/ZaporojetsDJDD22}, we apply a simple span proposal step to keep only the top-$k$ spans based on the span score $\Phi(s)$ from a feed-forward neural net (FFNN):
\[
\Phi(s) = \mathrm{FFNN}(\mathbf{x}_s).
\]
Then the generated $k$ candidate spans, tipped as argument spans most likely, will insert to the AMR graph $\mathcal{G}$ to construct our proposed tailored AMR graph $\mathcal{G}_t$. We analyze the influence of the choice of $k$ in Appendix-\ref{sec:choice-of-k} on the recall and efficiency.


We also minimize the following binary cross entropy loss to train the argument identification:
\[
\mathcal{L}_{span} = - (y \log(\Phi(\mathbf{x})) + (1 - y)\log(1 - \Phi(\mathbf{x}))),
\]
where $y$ is assigned the true label when the offsets of corresponding span match the golden-standard argument span, otherwise, the false label.

\paragraph{AMR Graph Module}
As introduced in Sec-\ref{sec:tAMR}, 
the embedding of each node $u_s$ in $\mathcal{G}_t$ is initialized by the aligned span representation $\mathbf{x}_s$ and its type embedding:
\[
\mathbf{g}^0_{u_s} = \mathrm{LayerNorm}(\mathbf{x}_s + \mathcal{T}_{node}(u_s)),
\]
where $\mathcal{T}_{node}$ refers to the lookup table about node types, composed of \{\textit{trigger, surrounding trigger, candidate span, others}\} four types. The newly inserted nodes are connected to their neighbor nodes, which are close in the text sequence, with a new edge type \texttt{context}.  

We use $L$-layer stacked R-GCN \citep{DBLP:conf/esws/SchlichtkrullKB18} to model the interactions among different nodes through edges with different relation types. The hidden states of nodes in $(l+1)^{th}$ layer can be formulated as:
\[
\mathbf{g}_u^{l+1}\!=\!\mathrm{ReLU}(\bW_0^{(l)}\mathbf{g}_u^{(l)}\!+\!\sum_{r\in R}\!\sum_{v\in N_{u}^r}\!\frac{1}{c_{u, r}}\!\bW_r^{(l)}\mathbf{g}_{v}^{(l)}),
\]
where $R$ is the clusters of AMR relation types in Table~\ref{tab:amr-edge-type}, $N_u^r$ denotes the set of neighbor nodes of node $u$ under relation $r \in R$ and $c_{u, r}$ is a normalization constant. $\bW_0^{(l)}, \bW_r^{(l)}$ are trainable parameters.

We concatenate hidden states of all layers and derive the final node representation $\mathbf{g}_u = \bW_g[g^0_u; g^1_u; \dots, g^L_u]$.

\paragraph{Classification Module}
We perform multi-class classification to predict what role a candidate span plays, or it does not serve as an argument.
As mentioned in Sec-\ref{sec:link-pred}, we take the node representation $\mathbf{g}_{u_s}$ and $\mathbf{g}_{u_\tau}$ which denote the aligned candidate span $s$ and trigger $\tau$, respectively. Following \citet{DBLP:conf/naacl/XuWLZCS22}, we also concatenate the event type embedding. The final classification representation can be formulated as:
\[
\mathbf{z}_{s} = [\mathbf{g}_{u_s}; \mathbf{g}_{u_\tau}; \mathcal{T}_{event}(e)].
\]

We adopt the cross entropy loss function:
\[
\mathcal{L}_c = -\sum_{s}y_s\log P(\hat{r}_s=r_s),
\]
where $\hat{r}_s$ is logits obtained by a FFNN on $\mathbf{z}_s$, and $r_s$ is the gold argument role of span $s$.

We train the model using the multi-task loss function $\mathcal{L} = \mathcal{L}_c + \lambda \mathcal{L}_s$ with hyperparameter $\lambda$. As a result, argument classification can be positively affected by argument identification.

\section{Experiments}

\subsection{Datasets and Evaluation Metrics}

We evaluate our model on two commonly used document-level event argument extraction datasets, WikiEvents \citep{DBLP:conf/naacl/LiJH21} and RAMS \citep{DBLP:conf/acl/EbnerXCRD20}. WikiEvents contains more than 3.9k samples, with 50 event types and 59 argument roles. RAMS is a benchmark that emphasizes the cross-sentence events, which has 9124 annotated events, containing 139 event types and 65 kinds of argument roles. We follow the official train/dev/test split for WikiEvents and RAMS, and leave the detailed data statistics in Appendix-\ref{sec:data-stat}.

For WikiEvents, we evaluate two subtasks of event argument extraction. \textbf{Arg Identification}: An argument span is correctly identified if the predicted span boundary match the golden one. \textbf{Arg Classification}: If the argument role also matches, we consider the argument is correctly classified. Following \citet{DBLP:conf/naacl/LiJH21}, we report two metrics, Head F1 and Coref F1. Head F1 measures the correctness of the head word of an argument span, the word that has the smallest arc distance to the root in the dependency tree. For Coref F1, the model is given full credit if the extracted argument is coreferential with the reference as used in \citet{DBLP:conf/acl/JiG08}. In addition, for RAMS dataset, we mainly concern Arg Classification and report the Span F1 and Head F1. For a sufficient comparison, We follow \citet{DBLP:conf/acl/MaW0LCWS22} and additionally evaluate Span F1 for Arg Identification on the test set.

\begin{table}[!tbp]
\centering
\small
\caption{\textbf{Main results on the WikiEvents test set.} Rows in gray are results of our proposed models that perform best on the development set, and subscripts denote the standard deviation computed from 3 runs. Models under the double line are based on large models with similar model sizes. The best results are in \textbf{bold} and the previous best results are \underline{underlined}.}  %
\resizebox{\columnwidth}{!}{
\begin{tabular}{lcccc}
    \toprule
    \multirow{2}*{\bf Model} & \multicolumn{2}{c}{\bf Arg Identification} & \multicolumn{2}{c}{\bf Arg Classification} \\
    \cmidrule(lr){2-3}\cmidrule(lr){4-5} & Head F1 & Coref F1 & Head F1 & Coref F1 \\
    \midrule
    \multicolumn{5}{l}{\textit{BERT-base}} \\
    BERT-CRF & 69.83 & 72.24 & 54.48 & 56.72 \\
    BERT-QA & 61.05 & 64.59 & 56.16 & 59.36 \\
    BERT-QA-Doc & 39.15 & 51.25 & 34.77 & 45.96 \\
    EEQA & - & - & 56.9 & - \\
    TSAR & \underline{75.52} & \underline{73.17} & \underline{68.11} & \underline{66.31} \\
    \rowcolor{gray!20} TARA & 76.49 & 74.44 & \bf 70.52 & 68.47 \\
    \rowcolor{gray!20} TARA$_\textrm{compress}$ & \bf 76.76 & \bf 74.88 & 70.18 & \bf 68.67 \\
    \midrule
    \midrule
    \multicolumn{5}{l}{\textit{BART-large}} \\
    BART-Gen & 71.75 & 72.29 & 64.57 & 65.11 \\
    PAIE & - & - & 68.4 & - \\
    EA$^2$E & 74.62 & \underline{75.77} & 68.61 & \underline{69.70} \\
    \midrule
    \multicolumn{5}{l}{\textit{RoBERTa-large}} \\
    EEQA & - & - & 59.3 & - \\
    TSAR & \underline{76.62} & 75.52 & \underline{69.70} & 68.79 \\
    \rowcolor{gray!20} TARA & \bf 78.64$_{0.16}$ & 76.40$_{0.23}$ & 72.89$_{0.27}$ & 70.95$_{0.23}$ \\
    \rowcolor{gray!20} TARA$_{\textrm{compress}}$ & 78.50$_{0.34} $ & \bf 76.71$_{0.14} $ & \bf 73.33$_{0.41}$ & \bf 71.55$_{0.25} $ \\
    \bottomrule
\end{tabular}}

\label{tab:wikievents}
\end{table}

\subsection{Settings}
We adopt the transition-based AMR parser proposed by \citet{DBLP:conf/emnlp/AstudilloBNBF20} to obtain the AMR graph with node-to-text alignment information, which can achieve satisfactory results for downstream tasks. We also show the performance using another state-of-the-art AMR parser, AMRBART \citep{DBLP:conf/acl/00010022}, in Appendix-\ref{sec:amr-parser}. Besides, we use BERT$_\textrm{base}$ and RoBERTa$_\textrm{large}$ provided by huggingface\footnote{https://huggingface.co/} as the backbone. The models are trained with same hyper-parameters as \citet{DBLP:conf/naacl/XuWLZCS22}, details listed in Appendix-\ref{sec:hparam}. Experiments based on base models are conducted on a single Tesla T4 GPU, and large models on 4 distributed Tesla T4 GPU in parallel.

\subsection{Main Results}
\label{sec:main-results}

\begin{table}[!tbp]
\centering
\small
\caption{\textbf{Main results on the RAMS dataset.} \textbf{Arg-I} denotes Span F1 for the Arg Identification subtask, and other metrics are for the Arg Classification subtask. * indicates results reported by \citet{DBLP:conf/acl/MaW0LCWS22}.}
\resizebox{\columnwidth}{!}{
\begin{tabular}{lccccc}
    \toprule
    \multirow{2}*{\bf Model} & \multicolumn{2}{c}{\bf Dev} & \multicolumn{3}{c}{\bf Test} \\
    \cmidrule(lr){2-3}\cmidrule(lr){4-6} & Span F1 & Head F1 & Span F1 & Head F1 & Arg-I \\
    \midrule
    \multicolumn{6}{l}{\textit{BERT-base}} \\
    BERT-CRF & 38.1 & 45.7 & 39.3 & 47.1 & - \\
    BERT-CRF$_\textrm{TCD}$ & 39.2 & 46.7 & 40.5 & 48.0 & -\\
    Two-Step & 38.9 & 46.4 & 40.1 & 47.7 & - \\
    Two-Step$_\textrm{TCD}$ & 40.3 & 48.0 & 41.8 & 49.7 & - \\
    FEAE & - & - & 47.40 & - & \bf \underline{53.49} \\
    TSAR & \underline{45.23} & \underline{51.70} & \bf \underline{48.06} & \underline{55.04} & - \\
    \rowcolor{gray!20} TARA & 45.81 & \bf 53.22 & \bf 48.06 & 55.23 & 52.82 \\
    \rowcolor{gray!20} TARA$_\textrm{compress}$ & \bf 45.89 & 53.15 & 47.43 & \bf 55.24 & 52.34 \\
    \midrule
    \midrule
    \multicolumn{6}{l}{\textit{BART-large}} \\
    BART-Gen & - & - & 48.64 & 57.32 & 51.2* \\
    PAIE & - & - & \underline{52.2} & - & \underline{56.8} \\
    \midrule
    \multicolumn{6}{l}{\textit{RoBERTa-large}} \\
    TSAR & \underline{49.23} & \underline{56.76} & 51.18 & \underline{58.53} & - \\
    \rowcolor{gray!20} TARA & 50.01$_{0.20}$ & 58.17$_{0.16}$ & \bf 52.51$_{0.05}$ & \bf 60.86$_{0.12}$ & \bf 57.11$_{0.10}$ \\
    \rowcolor{gray!20} TARA$_{\textrm{compress}}$ & \bf 50.33$_{0.17}$ & \bf 58.49$_{0.30}$ & 52.28$_{0.15}$ & 60.73$_{0.10}$ & 56.91$_{0.17}$ \\
    \bottomrule
\end{tabular}}

\label{tab:rams}
\end{table}

We compare our model with several baselines and the following previous state-of-the-art models. (1) QA-based models: \textbf{EEQA} \citep{DBLP:conf/emnlp/DuC20} and \textbf{FEAE} \citep{DBLP:conf/acl/WeiSZZGJ20}. (2) Generation-based models: \textbf{BART-gen} \citep{DBLP:conf/naacl/LiJH21}, \textbf{PAIE} \citep{DBLP:conf/acl/MaW0LCWS22}, and \textbf{EA$^2$E} \citep{DBLP:conf/naacl/ZengZJ22}. (3) Span-based models: \textbf{TSAR} \citep{DBLP:conf/naacl/XuWLZCS22}. TSAR is the first and sole work utilizing AMR for Doc-level EAE.

Table~\ref{tab:wikievents} illustrates the results on the WikiEvents test set. As is shown, our proposed methods consistently outperform previous works with different sized backbone models. TARA$_{\textrm{compress}}$ achieves comparable results with TARA, with more than 30\% nodes and edges being pruned, which suggests that the compression process is effective. We compare the better one with other models in the following analysis.

More than 4pt Head F1 for Arg Classification against approaches that do not use AMR indicates the value of deep semantic information. TSAR is the only work to introduce AMR to document-level EAE tasks, but utilizes AMR graphs in an implicit way of decomposing the node representations to contextual representations. The 3.63pt performance gain compared to TSAR shows that our method, which explicitly leverages AMR graphs to perform link prediction, can make better use of rich semantic structures provided by AMR. Besides, EA$^2$E learns event-event relations by augmenting the context with arguments of neighboring events, which may bring noises in the inference iteration, while we simply mark nodes of other event triggers in the graph and yields an improvement of 4.72pt Head F1.

Comparing the identification and classification scores, we find that the performance gain of the latter is always higher, which indicates that our method not only helps the model find more correct arguments but also increases the accuracy of classifying argument roles. Another finding is that our method contributes more to Head F1 instead of Coref F1 in most cases. The main difference between the two metrics is boundary correctness. The result suggests that although our method helps less in identifying the span boundary, it enhances the capability of finding arguments. Our model is less powerful in span boundary identification is reasonable since the span proposal module only takes the textual information, and we will consider upgrading the span proposal module with AMR information in future work.

Similar conclusion can be drawn from Table~\ref{tab:rams}\footnote{We did not mark surrounding events for RAMS due to the lack of annotations.}, which compares our method with previous works in both dev and test sets of RAMS. Our method achieves new state-of-the-art results using the large model with 2.33pt Head F1 improvement on the test set compared with TSAR, and yields comparable results based on BERT$_\textrm{base}$. PAIE manually creates a set of prompts containing event descriptions for each event type, providing additional knowledge which benefits most for classification with numerous classes. In contrast, our method improves up to 0.31/0.31pt Span F1 for Arg Identification/Classification with the help of explicit AMR information.

\section{Analysis}
\subsection{Ablation Study}
\label{sec:ablation}
\begin{table}[!tbp]
\centering
\small
\caption{Ablation study on the WikiEvents test set based on RoBERTa$_\textrm{large}$. ``wo'' denotes ``without''.}
\resizebox{\columnwidth}{!}{
\begin{tabular}{lcccc}
    \toprule
    \multirow{2}*{\bf Model} & \multicolumn{2}{c}{\bf Arg Identification} & \multicolumn{2}{c}{\bf Arg Classification} \\
    \cmidrule(lr){2-3}\cmidrule(lr){4-5} & Head F1 & Coref F1 & Head F1 & Coref F1 \\
    \midrule
    \bf TARA & \bf 78.64 & \bf 76.40 & \bf 72.89 & \bf 70.95 \\
    (a) wo AMR & 75.04 & 73.79 & 68.94 & 68.04 \\
    (b) implicit AMR & 76.34 & 73.98 & 70.00 & 68.36 \\
    (c) wo span proposal & 70.84 & 67.71 & 64.38 & 61.84 \\
    (d) wo surrounding events & 77.15 & 75.76 & 71.48 & 70.27 \\
    (e) homogeneous graph & 77.87 & 75.88 & 71.54 & 69.74 \\  
    (f) fully-connected graph & 76.95 & 75.30 & 70.52 & 69.42 \\  
    \bottomrule
\end{tabular}}

\label{tab:ablation}
\end{table}

We perform ablation study to explore the effectiveness of different modules in our proposed model. Table~\ref{tab:ablation} provides results on the WikiEvents test set based on RoBERTa$_{\textrm{large}}$ when excluding various modules at a time, which helps us answer the following three crucial questions:

\paragraph{What is the effect of explicit AMR graphs?} (a): When we throw away the whole AMR graph and depend solely on the contextual representations from PLM to extract arguments, the Head F1 of Arg Classification decreases by a large margin of 3.95pt, due to the lack of deep semantic information provided by AMR. Besides, (b): implicitly utilizing AMR by taking AMR edge classification as an auxiliary task, leads to a performance drop by 2.89pt. It suggests that explicitly using AMR graphs is more practical for document understanding and argument extraction.

\paragraph{What is the effect of tailored AMR graphs for EAE?} (c): Once we drop spans that are not aligned with an AMR node, there is a sharp decrease up to 8.51pt Head F1, demonstrating the necessity of span proposal. 
(d): If we do not mark surrounding event triggers in the AMR graph, the Head F1 gains a rise by 2.54pt compared to (a), but drops by 1.41pt compared to TARA using the unabridged tailored AMR graph, which shows that barely indicating surrounding events benefits to make full use of event-event relations.

\paragraph{What is the effect of heterogeneous graph structures?} (e): The removal of different edge types in the AMR graph, causes a slight performance drop by 1.35pt, illustrating the effectiveness of various edge types. In addition, (f): when we further remove the edge relations and replace the graph structure with a fully-connected layer, the performance decreases by 2.37pt. It suggests that the edge relations are also useful. Moreover, we find that (f) outperforms (a) with an improvement of 1.58pt Head F1, which indicates that the node-level representations are more expressive than word-level representations.

\subsection{Efficiency}
\begin{table}[!tbp]
\centering
\small
\caption{Time cost (in seconds) of different AMR-guided models based on RoBERTa$_\textrm{large}$ on the test set of WikiEvents. Experiments are run for one epoch on 4 same Tesla T4 GPUs.}
\begin{tabular}{lcc}
    \toprule
    \bf Model & \bf Training Time & \bf Inference Time \\
    \midrule
    TSAR & 603.52 & 33.43 \\
    TARA & 319.63 & 15.56 \\  %
    TARA$_\textrm{compress}$ & \bf 281.92 & \bf 14.70 \\
    \bottomrule
\end{tabular}
\label{tab:time}
\end{table}

\begin{figure*}[t]
    \centering
    \includegraphics[width=0.9\linewidth]{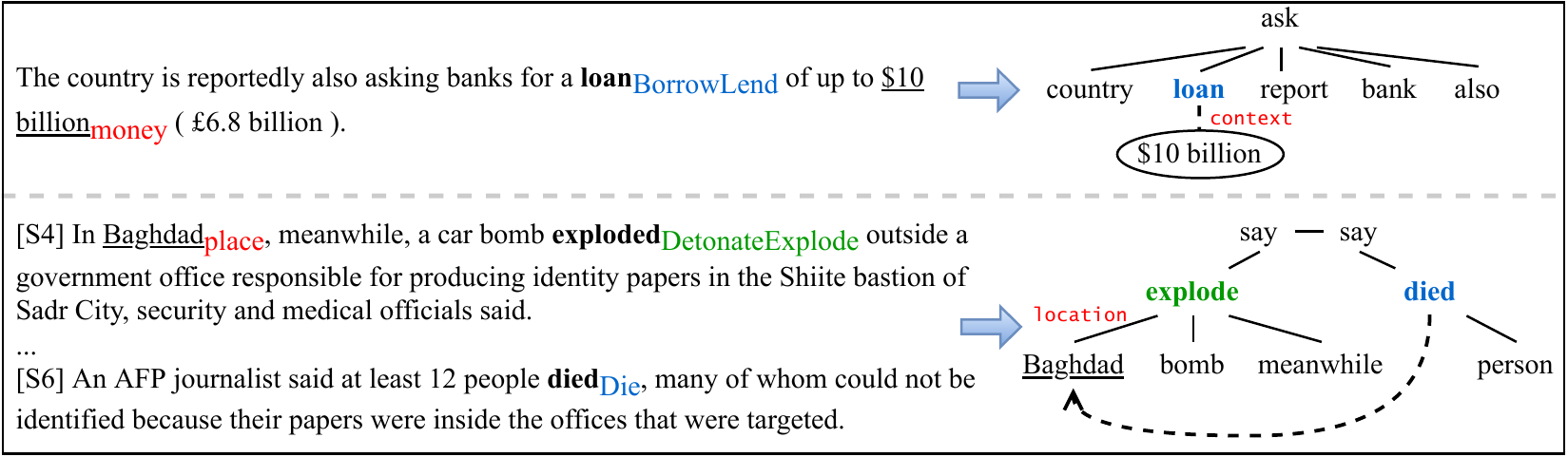}
    \caption{Case study for TAG. The left are excerpted text sequences, with \textbf{triggers} in bold, \underline{arguments} underlined, and event types and argument roles as subscripts. The right are corresponding AMR graphs.}  
    \label{fig:case-study}
\end{figure*}

Table~\ref{tab:time} reports the efficiency of different models using AMR graphs. TSAR encodes the input document from local and global perspectives and obtains AMR-enhanced representations by fusing contextual and node representations, while TARA directly utilize AMR graphs to perform link prediction. Though the two models share similar model sizes, TARA runs approximately 2 times faster than TSAR and saves up to 53\% inference time.
When compressing the AMR graph in the pre-processing stage, with more than 30\% nodes and edges omitted, TARA$_\textrm{compress}$ speeds up further, resulting in 56\% inference time saving.

\subsection{Error Analysis}
\label{sec:error-analysis}
\begin{table}[!tbp]
\centering
\small
\caption{Error analysis on the WikiEvents test set based on RoBERTa$_\textrm{large}$.}
\resizebox{\columnwidth}{!}{
\begin{tabular}{lcccc}
    \toprule
    \bf Model & \bf Missing Head & \bf Overpred Head & \bf Wrong Role & \bf Wrong Span \\
    \midrule
    Baseline & 137 & 110 & 37 & 18 \\
    TSAR & 136 & 104 & 34 & 20 \\
    TARA & 120 & 98 & 33 & 19 \\
    \bottomrule
\end{tabular}}

\label{tab:error}
\end{table}

To compare different models in greater detail, we explore four specific error types listed in Table~\ref{tab:error}. \textit{Missing Head} refers to the number of missing arguments, those the model dose not predict that they are arguments, and we only consider the head word to relax the boundary constraint. \textit{Overpred Head} denotes the number of spans that the model falsely assumes that they play a role. Besides, even though the model succeeds to identify the head word of a golden span, it can still assign wrong argument role to it, which we call \textit{Wrong Role}; or it cannot predict the true start and end position of the span, named \textit{Wrong Span}. We suppose extracting coreferential arguments is reasonable. Baseline refers to (a) in Sec-\ref{sec:ablation}, which performs worse than TSAR and TARA.

As shown in Table~\ref{tab:error}, TARA misses fewer argument spans compared to TSAR. In addition, while finding more correct argument spans, TARA dose not predict more wrong roles, that is, it will improve more on the Arg Classification subtask. The first three error types are usually attributed to the severe class imbalance problem. With few examples of one label, the model cannot successfully learn the meaning of it and thus is hard to assign it to the true arguments. Moreover, our proposed model does not do better in recognizing correct span boundary considering \textit{Wrong Span}. We observe that most Wrong Span errors result from the inconsistency of the annotation in the dataset, e.g., whether articles (such as \textit{the} and \textit{a}), adjectives and quantifiers before a noun should be included to a span.

\subsection{Case Study for TAG}

In this section, we look into specific examples to explore how tailored AMR graphs work. Firstly, the top part of Figure~\ref{fig:case-study} 
illustrates the effect of adding missing spans to AMR graphs. Though AMR compresses the text sequence to a deep semantic structure, it may have a different focus from event structures. For instance, ``\textit{\$10 billion}'', which plays an argument role of \textit{Money} for event \textit{BorrowLend}, is left out by the vanilla AMR graph. In contrast, TAG will add the span and successfully serve for link prediction. Additionally, as shown in the bottom part of the figure, there are two events share the same argument ``\textit{Baghdad}'', and Baseline can not correctly identify the argument for the further event ``\textit{died}'' while TARA does both right. That is because when indicating surrounding event triggers in the graph, the event ``\textit{died}'' would pay attention to the subgraph of the event ``\textit{explode}'' and identify the implicit argument through a closer path in the graph than in the text sequence.


\section{Related Work}

Doc-level EAE is a frontier direction of Event Extraction and has received broad attention from industry and academia in recent years. Unlike the well-developed sentence-level event extraction \citep{DBLP:conf/acl/Xi0ZWJW20, DBLP:conf/emnlp/MaWABA20}, the Doc-level EAE faces more challenges. \citet{DBLP:conf/naacl/LiJH21} proposes an end-to-end generation-based approach for Doc-level EAE. \citet{DBLP:conf/ijcai/FanWLZSH22} and \citet{DBLP:conf/acl/XuLLC20} construct an entity-based graph to model dependencies among the document. \citet{DBLP:conf/acl/DuC20} chooses the hierarchical method to aggregate information from different granularity. 

Recently, there has been a rising trend of utilizing AMR information to assist event extraction. \citet{DBLP:conf/naacl/XuWLZCS22} employs node representations derived by AMR graphs. \citet{DBLP:conf/ijcai/LinC0J022} and \citet{DBLP:journals/corr/abs-2205-12490} introduce AMR path information as training signals to correct argument predictions. \citet{DBLP:conf/acl/WangW0L000L020} pre-trains the EAE model with a contrastive loss built on AMR graphs. However, previous works have only treated AMR as an auxiliary feature or supervised signal and has not fully exploited the correlation between AMR and EAE. As the scheme of the AMR graph is very similar to the event structure (predicate-arguments vs. trigger-arguments), EAE can be reformulated as an AMR-based problem. With TAG, we can define EAE as a task only related to graphs and conditionally independent of documents, thus achieving a simpler and more efficient model.

Previous works also explore the ways of enriching AMR graphs to suit information extraction tasks. \citet{DBLP:conf/ijcai/FanWLZSH22} trains a learnable module to add nodes and edges to the AMR graph. \citet{DBLP:conf/naacl/ZhangJ21} discusses different ways to integrate missing words with the AMR graph. While these methods tend to enlarge AMR graphs, causing a larger graph size and increasing the training difficulty, our method compresses the irrelevant information in AMR to improve efficiency and help the model to be concentrated.

\section{Conclusion}
We propose to reformulate document-level event argument extraction as a link prediction problem on our proposed tailored AMR graphs. With adding missing spans, marking surrounding events, and removing noises, AMR graphs are tailored to EAE tasks. We also introduce a link prediction model based on TAG to implement EAE. Elaborate experiments show that explicitly using AMR graphs is beneficial for argument extraction.

\section*{Limitations}
Firstly, as analyzed in Sec-\ref{sec:error-analysis}, our proposed method fails to make a significant improvement on span boundary identification. For one thing, the annotation inconsistency in the dataset hinders the model's understanding. For another, our span proposal module leverages the contextual information alone with implicit training signals for span boundary information. We will consider enhancing the span proposal module with AMR information in the future. Secondly, though TARA saves up to 56\% inference time compared to the previous AMR-guided work, its entire training requires more than 7h on 4 Tesla T4 GPUs. The bottleneck is the incongruity of pre-trained language models and non-pre-trained GNNs. We leave the problem for future work. Finally, arguments on Wikievents and RAMS are still relatively close to its event trigger (e.g., RAMS limits the scope of arguments in a 5-sentence window), and thus connecting sentence-level AMR graphs is enough to model the long-distance dependency. Otherwise, document-level AMR graphs with coreference resolution are in demand.

\section*{Ethics Statement}
Our work complies with the ACL Ethics Policy. As document-level event argument extraction is a standard task in NLP, we do not see any critical ethical considerations. We confirm that the scientific artifacts used in this paper comply with their license and intended use. Licenses are listed in Table~\ref{tab:license}.

\section*{Acknowledgement}
We would like to express our sincere gratitude to the reviewers for their thoughtful and valuable feedback. This work was supported by the National Key Research and Development Program of China (No.2020AAA0106700) and National Natural Science Foundation of China (No.62022027).

\bibliography{anthology,custom}
\bibliographystyle{acl_natbib}

\clearpage
\appendix

\section{Appendix}
\label{sec:appendix}

\begin{table}[!tbp]
\centering
\small
\caption{Licenses of scientific artifacts used in this paper.}
\begin{tabular}{lc}
    \toprule
    \bf Scientific Artifact & \bf License \\
    \midrule    
    WikiEvents & MIT License \\
    RAMS & Apache License 2.0 \\
    bert-base-uncased & Apache License 2.0 \\
    roberta-large & MIT License \\
    \bottomrule
\end{tabular}
\label{tab:license}
\end{table}

\subsection{Statistics of Datasets}
\label{sec:data-stat}

\begin{table}[!tbp]
\centering
\small
\caption{Statistics of WikiEvents and RAMS datasets.}
\begin{tabular}{lcccc}
    \toprule
    \bf Dataset & \bf Split & \bf \#Docs & \bf \#Events & \bf \#Arguments \\
    \midrule    
    \multirow{3}*{WikiEvents} & Train & 206 & 3,241 & 4,542 \\
     & Dev & 20 & 345 & 428 \\
     & Test & 20 & 365 & 566 \\
    \midrule
    \multirow{3}*{RAMS} & Train & 3,194 & 7,329 & 17,026 \\
     & Dev & 399 & 924 & 2,188 \\
     & Test & 400 & 871 & 2,023 \\
    \bottomrule
\end{tabular}
\label{tab:data-stat}
\end{table}


The details of statistics of WikiEvents and RAMS datasets are listed in Table~\ref{tab:data-stat}.

\subsection{Hyperparameters}
\label{sec:hparam}

We set batch size to 8, and train the model using AdamW \citep{DBLP:conf/iclr/LoshchilovH19} optimizer and a linearly decaying scheduler \citep{DBLP:journals/corr/GoyalDGNWKTJH17} with 3e-5 learning rate for pre-trained language encoders and 1e-4 for other modules. For Wikievents, we train the model for 100 epochs, and set $\lambda$ to 1.0 and $L$ to 3. For RAMS, we train the model for 50 epochs, and set $\lambda$ to 0.05 and $L$ to 4.

\subsection{The choice of $k$}

\label{sec:choice-of-k}
\begin{figure}[!tbp]
    \centering
    \includegraphics[width=\linewidth]{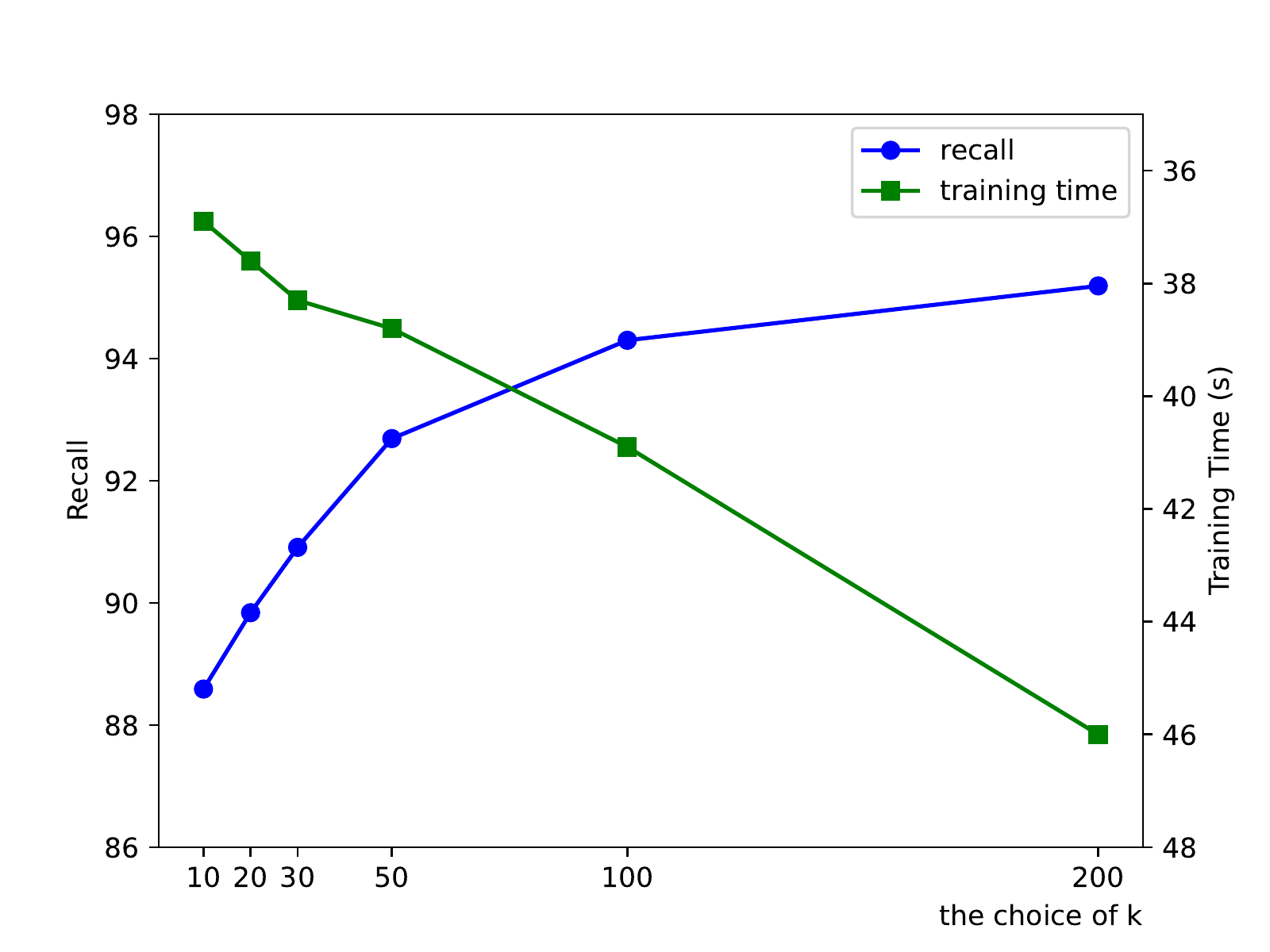}
    \caption{Recall and Training Time with respect to the choice of $k$. Experiments are run for one epoch on a single Tesla T4 GPU.}
    \label{fig:exp-k}
\end{figure}
\begin{table}[!tbp]
\centering
\small
\caption{Comparison between transition-based AMR parser (abbrev. transition-AMR) and AMRBART on WikiEvents test set based on RoBERTa$_\textrm{large}$.}
\resizebox{\columnwidth}{!}{
\begin{tabular}{lccccc}
    \toprule
    \multirow{2}*{\bf Model} & \bf AMR 2.0 & \multicolumn{2}{c}{\bf Arg Identification} & \multicolumn{2}{c}{\bf Arg Classification} \\
    \cmidrule(lr){2-2}\cmidrule(lr){3-4}\cmidrule(lr){5-6} & Smatch & Head F1 & Coref F1 & Head F1 & Coref F1 \\
    \midrule
    transition-AMR & 81.3 & \bf 78.64 & 76.40 & 72.89 & 70.95 \\
    transition-AMR$_{\textrm{compress}}$ & 81.3 & 78.50 & \bf 76.71 & \bf 73.33 & \bf 71.55 \\
    AMRBART & \bf 85.4 & 78.35 & 76.29 & 73.07 & 70.83 \\  
    \bottomrule
\end{tabular}}
\label{tab:amr-parser}
\end{table}

Span proposal module is of great importance to construct the tailored AMR graph, and intuitively, selecting different number of spans as candidates for Arg Classification will exert an influence on performance and efficiency. Therefore, we present visually the trend of recall and inference time when ranging $k$, which denotes the number of proposed spans. As illustrated in Figure~\ref{fig:exp-k}, as $k$ becomes larger, recall is higher, while inference is lower. Moreover, when recall of span proposal is low, a number of positive examples for Arg Classification would be dropped, which impedes the model to learn argument roles. On the other hand, too many candidate spans aggravate the problem of class imbalance. As a consequence, we make a trade-off to set $k = 50$.

\subsection{AMR parsers}
\label{sec:amr-parser}

TARA, as the name implies, relies on automatic AMR parsers to build signals of message passing. To explore the effect of different AMR parsing performance, we compare test results of TARA using transition-based AMR parser and a latest state-of-the-art parser AMRBART \citep{DBLP:conf/acl/00010022} in Table~\ref{tab:amr-parser}. We implement a simple node-to-text aligner and compress the obtained AMR graph as described in Sec-\ref{sec:compress} for AMRBART. As shown in the table, though AMRBART brings better AMR parsing performance, it dose not gain more improvements for EAE. It demonstrates that there is still a gap between AMR graphs and event structures. Nonetheless, TARA equipped with AMRBART consistently outperforms previous models, which indicates the robustness of our proposed model.

\section{Subgraph Compression}
\label{sec:compress}
\begin{figure}[t]
    \centering
    \includegraphics[width=0.8\linewidth]{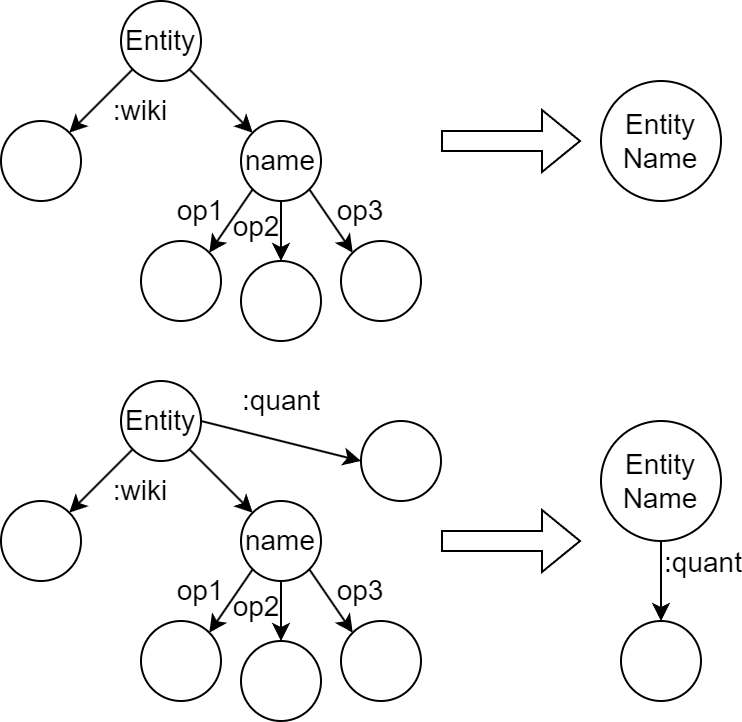}
    \caption{Compression rules. See text for details.}
    \label{fig:app_compress}
\end{figure}
As mentioned in the main text, we compress the subgraph to make the graph compact. Figure \ref{fig:app_compress} illustrates how we compress a subgraph. Firstly, we will find a subgraph that has an AMR label in pre-defined entity types. The type list is induced from the AMR parser configurations, and we also give the list here, \textit{Country, Quantity, Organization, Date-attrs, Nationality, Location, Entity, Misc, Ordinal-entity, Ideology, Religion, State-or-province, Cause-of-death, Title, Date, Number, Handle, Score-entity, Duration, Ordinal, Money, Criminal-charge, Person, Thing, State, Date-entity, Name, Publication, Province, Government-organization, City-district, City, Criminal-organization, Group, Religious-group, String-entity, Political-party, World-region, Country-region, String-name, URL-entity, Festival, Company, Broadcast-program}. If such a node has a child node with the label ``name'' and outgoing edges like ``op1'', ``op2'', we will compress this subgraph. The compression merges labels of all nodes connected with ``op1'', ``op2'', ``op3'' edges as a phrase according to the ascending order of edges. The text alignment information of the merged node becomes the range from the most left position to the most right position of nodes in the subgraph, which means there is a little chance to enlarge the corresponding text span if the original positions are discontinuous. The compression will preserve all incoming and outgoing edges except the edge ``:wiki''. As shown in the Figure \ref{fig:app_compress}, we keep the ``:quant'' edge but remove the ``:wiki'' edge.

\end{document}